# Robustness-Driven Exploration with Probabilistic Metric Temporal Logic

Xiaotian Liu
Computer Science Department
Wake Forest University
Winston-Salem, NC, USA
liux16@wfu.edu

Pengyi Shi
Computer Science Department
Wake Forest University
Winston-Salem, NC, USA
ship16@wfu.edu

Sarra Alqahtani
Computer Science Department
Wake Forest University
Winston-Salem, NC, USA
sarra-alqahtani@wfu.edu

Victor Pauca
Computer Science Department
Wake Forest University
Winston-Salem, NC, USA
paucavp@wfu.edu

Miles Silman
Biology Department
Wake Forest University
Winston-Salem, NC, USA
silmanmr@wfu.edu

## ABSTRACT

The ability to perform autonomous exploration is essential for unmanned aerial vehicles (UAV) operating in unstructured or unknown environments where it is hard or even impossible to describe the environment beforehand. However, algorithms for autonomous exploration often focus on optimizing time and coverage in a greedy fashion. That type of exploration can collect irrelevant data and wastes time navigating areas with no important information. In this paper, we propose a method for exploiting the discovered knowledge about the environment while exploring it by relying on a theory of robustness based on Probabilistic Metric Temporal Logic (P-MTL) as applied to offline verification and online control of hybrid systems. By maximizing the satisfaction of the predefined P-MTL specifications of the exploration problem, the robustness values guide the UAV towards areas with more interesting information to gain. We use Markov Chain Monte Carlo to solve the P-MTL constraints. We demonstrate the effectiveness of the proposed approach by simulating autonomous exploration over Amazonian rainforest where our approach is used to detect areas occupied by illegal Artisanal Small-scale Gold Mining (ASGM) activities. The results show that our approach outperform a greedy exploration approach (Autonomous Exploration Planner) by 38% in terms of ASGM coverage.

## 1 INTRODUCTION

Exploration is often an important first step in tasks of robotics and autonomous vehicles, such as mapping, rescue missions, or path planning in unknown environments. Techniques that tackle this problem typically focus on exploration time and coverage, i.e. how fast and how much of an unexplored area can be explored [1-3]. Although optimizing coverage and time for exploration problems is crucial, it is important in some problem domains to consider exploiting the detected information about the environment while exploring it to prioritizing the exploration of interesting areas encountered during flight. Adding such spatial and temporal considerations into exploration enhances the decision robustness about the navigation behavior of the vehicle and introduces some predictability on where the vehicle could move next. Moreover, it is usually more desirable to gather knowledge and information about certain areas than wasting the vehicle's resources such as flight time or its local storage exploring the whole environment.

In this paper, we develop a novel robustness-driven exploration (RDE) approach to constrain a UAV movement according to user-defined spatial and temporal constraints expressed in a probabilistic extension of Metric Temporal Logic (P-MTL). These constraints guide the exploration into specific areas in the environment that we call Areas of Interest (AoI) based on the online detection of those areas. The first contribution of our work is the proposal of a method to explore unknown environment according to a robustness function that considers the degree of satisfaction of P-MTL specifications of AoI. By utilizing the notion of robustness for Metric Temporal Logic (MTL) [4], we can quantify how robustly a UAV's exploration decision satisfies a P-MTL specification. Large positive values suggest that the decision is robustly correct, while negative values imply falsification of the specification.

The second contribution of our work is adopting a Markov Chain Monte-Carlo (MCMC) sampling technique to solve the P-MTL constraints. Our approach performs a stochastic walk over the neighbor positions from the current position of the UAV guided by a robustness metric defined by the P-MTL. The MCMC technique is used as a local exploration strategy and is combined with a simplified version of Frontier Exploration [2] for global exploration. When a new AoI is available close to the UAV, the local exploration strategy is used, but when it is far away from any AoI, previously-seen but not-visited-yet positions with potential high robustness are explored instead. This simple technique helps the MCMC avoid getting stuck locally when exploring large areas with small or zero robustness values.

The performance of RDE is evaluated by simulating UAV exploration over the Amazon Forest in Peru to detect areas occupied by illegal mining (ASGM) activities. We test RDE against the Autonomous Exploration Planner (AEP) proposed in [3]. The results show that our proposed approach outperforms AEP in terms of AoI exploration by 38%.

The remainder of this paper is organized as follows. The next section further discusses the related work of the autonomous exploration and the temporal logic robustness and its application to exploration and navigation problems. In section 3, we introduce the problem definition and briefly review MTL robustness and P-MTL. Section 4 discusses the proposed approach, and Section 5 presents the results and discusses future work.

## 2 RELATED WORK

Autonomous exploration is a problem that has been intensively studied in the past three decades. The rise of UAV technology has increased the interest for this research area while also introducing many new challenges, especially when aiming for real-time exploration.

Early methods explored simple environments, for example, by following walls or similar obstacles. Frontier exploration [2] was the first exploration method that could explore a generic 2D environment. It defines frontier regions as the borders between free and unexplored areas. Exploration is done by sequentially navigating close frontiers. Repetition of this process leads to exploring the whole space. Advanced variants of this algorithm were presented in [5-7] also improving the coverage of unknown space along the path to the frontier.

Next-best-view (NBV) exploration is a common alternative to frontier-based exploration. A Receding Horizon NBV planner is developed in [1], for online autonomous exploration of unknown 3D spaces. The proposed planner employed the rapidly exploring random tree RRT with a cost function that considers the information gain at each node of the tree. A path to the best node was extracted and the algorithm was repeated after each time the vehicle moved along the first edge of the best path. An extension of this work is proposed in [3] to resolve the problem of getting stuck in local minima by extending it with frontier based planner for global exploration. Our approach also samples NBV according to the current vision of the UAV. In contrast to previously mentioned research, the views are randomly sampled as potential targets in our approach via MCMC and evaluated by their robustness values of the P-MTL constraints. In most cases, very few sampled positions suffice to determine a reasonably good next target.

Recently, temporal logics have been used in the context of robotic motion and path planning in unknown environments. For instance, deterministic μ-calculus was used to define specifications for sampling-based algorithms [8], Linear Temporal Logic (LTL) was coupled with RRT* [9], robustness of Metric Temporal Logic (MTL) has been embedded in A* [10] to increase the safety of UAVs navigating adversarial environments. Ayala et al. assumed that some properties of unknown environments can be identified earlier and used in Linear Temporal Logic (LTL) formulas, such that the exploration terminates once the formula is satisfied [11]. In [12], the researchers use co-safe LTL (cs-LTL) in their motion planning algorithm to compromise between satisfaction of customer demands and violation of road rules.

The closest work to our research is developed in [13]. The proposed approach includes the user-defined specification in the exploration problem using the safety fragment of the Signal Temporal Logic (STL). They embedded the robustness degree of STL specifications into the cost function of [3]. In contrast to their approach, we don't limit the user to certain fragments of the temporal logic. Instead, we propose to guide the exploration towards maximizing the robustness satisfaction of a complex specification defined using MTL and P-MTL.

## 3 PRELIMINARIES

In this section, we provide a formal definition of the problem we address in this research. Then, we introduce the syntax and semantics for MTL and P-MTL specifications and how we use them to formally define our exploration problem.

### 3.1 Problem Definition

We investigate the problem of exploring an unknown environment $E \subset \mathbb{R}^2$ by a UAV looking for certain areas of interests. Let the environment be represented by an occupancy map $M$ dividing $E$ into squared areas $m \in M$, that can be marked as Area of Interest (AoI) and non-AoI. The goal here is to increase the exploration of the areas that can be classified as AoI $E_{AoI} \subset E$ while decreasing the explored non-AoI, $E_{non-AoI} \subset E$, as much as possible. Such problem is considered solved when the UAV explored $E_{AoI} \backslash E_{res}$, where $E_{res}$ is a residual AoI left unexplored due to sensor detection limitations or UAV limitations. Object detection limitations include false negative classification of AoI as none-AoI and the UAV limitations may include insufficient battery life to explore all areas classified as AoI in $M$. The question we address in this work is: *which decisions should the UAV perform to explore $E_{AoI}$ completely and as fast as possible guided by the detection results of AoI*?

**Example (ASGM).** As a motivating example, we will consider the problem of mapping mercury-based ASGM in Amazonian Forest [14]. Mercury-based ASGM causes more mercury pollution than any other human activity on Earth, leading to major effects on the environment, health, and local economies. It is a global issue affecting 10 to 19 million people in over 70 countries [15]. Though satellite remote sensing would be ideal for monitoring ASGM sites (e.g. [16]), satellites do not currently produce images of sufficient resolution to accurately detect ASGM and differentiate, for example, between active and inactive mining sites (e.g. [17]). Moreover, satellite monitoring is not possible in cloudy and rainy weather which is very common in areas like the Amazon forest. UAVs can overcome those issues. They are affordable, easy to use, versatile, and even suitable in barely accessible areas. They also deliver high resolution data, mostly independent of cloud cover condition. However, the UAV field of view is significantly smaller than that of a satellite, and hence to cover a similar area efficiently; the acquisition of image data with UAV needs to be targeted to reduce the flight time, the required storage, and classification burden of the acquired images. According to researchers who collect images using a small UAV in Amazon Forest for their ASGM research [18], a full exploration of an area of 8x8 $km^2$ requires almost 8 separate flights; each flight has been done in 4 hours; acquiring 7200 images.

In this instance of the exploration problem, the UAV would explore the given bounded map looking for ASGM. The environment is unknown and the only input to the system is the onboard recognition system for materials and machinery used in ASGM. The goal for the UAV is to prioritize reaching the ASGM areas in its exploration while reducing the exploration time to better utilize the battery life.

The reachability requirement as described above is not enough to specify all system behaviors in practice [19]. This is especially true for autonomous vehicles wherein richer properties such as timing constraints, sequencing of events, conditional constraints, and the vehicle's safety are equally important. MTL is a popular formalism that can express such properties [4]. It can be used to assess not only whether a specific property is satisfied or not, but also helps to assess how much it is satisfied or not through its quantitative semantics [20], called MTL robustness. Since the object detection results of image data provided by the UAV sensors are usually in probabilistic representation, we utilize the robustness of P-MTL specifications in the context of UAV exploration behavior.

**PROBLEM (P-MTL Satisfaction).** For an P-MTL specification $\phi$, the P-MTL satisfaction problem consists of finding an output state $y$ of the system starting from some initial state $s_0 \in S$ under a control input signal $u \in U$ such that $y$ does satisfy the specification $\phi$ with the required robustness $\rho$.

An overview of our proposed approach to resolve Problem (P-MTL Satisfaction) is shown in Figure 1. At every time step, the object recognition module would generate probabilities for the detected AoI inside the UAV's vision. Then, based on the detection results, a Markov Chain Monte Carlo (MCMC) sampler would select a position $s$ from the set of neighbors and a vector of parameters that characterize the control input signal $u$ (i.e. speed and altitude). The selected position is then analyzed by the P-MTL robustness analyzer which would return a robustness score $\varepsilon$. In turn, if $\varepsilon$ is less than a predefined threshold $\rho$ then the stochastic sampler would be called again to select another position for analysis. If in this process, a position with $\varepsilon$ greater than $\rho$ is found, it is used by the path planner RA* [10] to move the UAV to that position. RA* has been originally implemented to embed MTL robustness into A* to avoid mobile obstacles in hostile environments. We change its MTL constraints to allow the UAV to explore the AoI available around the path while still be target oriented.

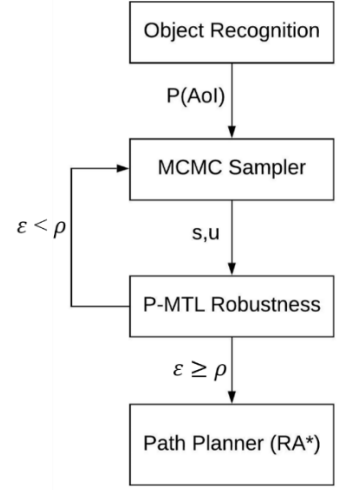

Figure 1: Proposed approach for RDE problem

### 3.2 MTL Robustness

**Defination1 (MTL Syntax).** Let AP be the set of atomic propositions and $I$ be a time interval of $\mathbb{R}$. The MTL $\varphi$ formula is recursively defined using the following grammar [4]:

$$\varphi \coloneqq T|p|\neg\varphi|\ \varphi_1 \vee \varphi_2 | \varphi_1 \wedge \varphi_2 |\ \varphi_1 \mathcal{U}_I \varphi_2 \qquad (1)$$

$T$ is the Boolean True, $p \in AP$, $\neg$ is the Boolean negation, $\vee$ and $\wedge$ are the logical OR and AND operators, respectively. $\mathcal{U}_I$ is the timed until operator and the interval $I$ imposes timing constraints on the operator. Informally, $\varphi_1 \mathcal{U}_I \varphi_2$ means that $\varphi_1$ must hold until $\varphi_2$ holds, which must happen within the interval $I$. The implication ($\Rightarrow$), Always ($\square$), Next ($\circ$), and Eventually ($\Diamond$) operators can be derived using the above operators.

To precisely capture the MTL formula, each predicate $p \in AP$ is mapped to a subset of the metric space $S$. Let $\mathcal{O}: AP \rightarrow \mathcal{P}(S)$ be an observation map for the atomic propositions. The Boolean truth value of a formula $\varphi$ with respect to the trajectory position $s$ at time $t$ is defined recursively using the MTL semantics directly reproduced as stated in [20]:

$(s, t) \coloneqq T \Leftrightarrow T$

$\forall\, p \in AP, (s, t) \coloneqq \mathcal{O}\, p \Leftrightarrow s_t \in \mathcal{O}(p)$

$(s, t) \coloneqq \mathcal{O}\, \neg \varphi \Leftrightarrow \neg(s, t) \coloneqq \mathcal{O}\, \varphi$

$(s, t) \coloneqq \mathcal{O}\, \varphi_1 \vee \varphi_2 \Leftrightarrow (s, t) \coloneqq \mathcal{O}\, \varphi_1 \vee (s, t) \coloneqq \mathcal{O}\, \varphi_2$

$(s, t) \coloneqq \mathcal{O}\, \varphi_1 \wedge \varphi_2 \Leftrightarrow (s, t) \coloneqq \mathcal{O}\, \varphi_1 \wedge (s, t) \coloneqq \mathcal{O}\, \varphi_2$

$(s, t) \coloneqq \mathcal{O}\, \varphi_1 \mathcal{U}_I \varphi_2 \Leftrightarrow \exists t' \in t + I.(s, t') \coloneqq \mathcal{O}\, \varphi_2\ \wedge \forall t'' \in (t, t'), (s, t'') \coloneqq \mathcal{O}\, \varphi_1$

In our problem of RDE, we have three atomic propositions including current battery level, battery minimum threshold, and AoI. To properly use the observation map semantics in the problem domain, we compute our propositions in terms of distance metric $d$, which is the Euclidian distance metric as described in [21]. We will provide more details about $d$ in the next section.

To formally measure the robustness degree of $\varphi$ at the trajectory position $s$ at time $t$, the robustness semantics of $\varphi$ is recursively defined as taken directly from [21]:

$$[\![T]\!](s,t) := +\infty$$
$$[\![p]\!](s,t) := Dist_d(s(t), \mathcal{O}(p))$$
$$[\![\neg\varphi]\!](s,t) := \neg[\![\varphi]\!](s,t)$$
$$[\![\varphi_1 \vee \varphi_2]\!](s,t) := [\![\varphi_1]\!](s,t) \sqcup [\![\varphi_2]\!](s,t)$$
$$[\![\varphi_1 \wedge \varphi_2]\!](s,t) := [\![\varphi_1]\!](s,t) \sqcap [\![\varphi_2]\!](s,t)$$
$$[\![\varphi_1 \mathcal{U}_{[l,u]} \varphi_2]\!](s,t) := \bigsqcup_{j=t+l}^{t+u} ([\![\varphi_2]\!](s,j) \sqcap \bigsqcap_{k=t}^{t-1} [\![\varphi_1]\!](s,k))$$

where $\sqcup$ stands for maximum, $\sqcap$ stands for minimum, $p \in AP$, and $l, u \in N$. The robustness is a real-valued function of the trajectory position s with the following important property stated in Theorem 1.

**Theorem 1**[21]: For any $s \in S$ and MTL formula $\varphi$, if $[\![\varphi]\!](s,i)$ is negative, then $s$ does not satisfy the specification $\varphi$ at time $i$. If it is positive, then $s$ satisfies $\varphi$ at $i$. If the result is zero, then the satisfaction is undefined.

MTL robustness is adopted in this research to measure how robust the exploration decision of the UAV at any point of time with respect to its specification expressed in MTL [21]. If an MTL specification $\varphi$ valuates to positive robustness $\varepsilon$, then the decision is right and, moreover, can tolerate perturbations up to $\varepsilon$ and still satisfy the specification. Similarly, if $\varepsilon$ is negative, then the decision does not satisfy $\varphi$ with a violation equal to $-\varepsilon$.

### 3.3 P-MTL

Probabilistic-MTL (P-MTL) [22] is an extension of MTL supporting reasoning over both stochastic states and stochastic predictions of states. The predictive operator $\bullet(t'|t)$ is used to refer to observed, estimated, and predicted states. The predictive operator is informally defined as follows:

Observed state value: $\bullet_t\, s$
Estimated state value: $\bullet_{t|t}\, s$
Predicted state value: $\bullet_{t'|t}\, s$

where $t$ is the observation time, $t'$ is the prediction time, and $s$ is the stochastic state under investigation. The value of $\bullet_t\, s$ is the observed value of state $s$ at time $t$. On the other hand, $\bullet_{t|t}\, s$ is the estimated value of state s at time $t$ which is the prediction made at time $t$ about the value of s at time $t$. This operator is useful when the detection results are in form of probability distribution. The value of $\bullet_{t'|t}\, s$ is a prediction made at time $t$ about the value of state s at time $t'$. $t'$ may be larger than $t$ (prediction about the future) or smaller than $t$ (prediction about the past).

## 4  AUTONOMOUS EXPLORATION WITH P-MTL ROBUSTNESS

Using the MTL syntax (*Definition 1*) and the informal definition of P-MTL, we define the P-MTL specification of our problem of RDE as follows.

$$\varphi = \Box(B > B_{min}) \wedge \Diamond\, p > \beta \wedge \Diamond\, p\left(\text{inside}(\bullet_{t-v|t}\, AoI)\right) < \lambda \Rightarrow \circ$$
$$p\left(\text{inside}\left(\bullet_{t|t}\, AoI\right)\right) > \lambda \qquad (2)$$

The first property of the formula $\Box(B > B_{min})$ represents a safety constraint requiring the UAV to keep its battery level above a certain threshold $B_{min}$ to get back to homebase. The threshold $B_{min}$ would be updated dynamically based on the current position of the UAV in the map. $\Diamond\, P\left(\text{inside}(\bullet_{t|t}\, AoI)\right) > \beta$ specifies that the UAV should be inside areas with a likelihood of being AoI above $\beta$. This property is classified as a liveness (i.e. preferred) property. To decrease the possibility that the UAV wastes time exploring non-AoI, the conditional liveness property $\Diamond\, P\left(\text{inside}(\bullet_{t-v|t}\, AoI)\right) < \lambda \Rightarrow \circ\, p\left(\text{inside}(\bullet_{t|t}\, AoI)\right) > \lambda$ asks the UAV to stay a maximum of $v$ time steps inside areas with likelihood of being $AoI$ less than $\lambda$ and when that happened the UAV must immediately (i.e. $\circ$ next decision) find another area with higher estimation of AoI or terminate the mission and go back to homebase.

In order to compute the P-MTL robustness for our exploration problem, we must define the Signed Distance, $Dist_d$ to reflect the domain properties [21]. In this paper, we define two functions to measure the distance from the propositions of the AoI and the minimum battery level B$_{min}$ (Figure 2). The location pins symbol represents AoI and the quadrotor drone symbol is the exploring UAV. The proposed approach analyzes the $n$ neighbor positions of the current position $s$ of the UAV and makes a decision about the next target based on the distance and depth functions given in Definition 2 and 3 next.

**Definition 2** (AoI *dist* function): Given that $s_i$ is the position that is under robustness analysis, $p$ is the probability of $s_i$ being inside an AoI given by the object recognition system, and $\beta$ is the minimum threshold for the detection results, the $dist_{AoI}$ between $s_i$ and the closest AoI is defined as

$$dist_{AoI}(s_i) = \begin{cases} (p - \beta) * 100 & if\ s_i \in \mathcal{O}(\beta) \\ 0 & otherwise \end{cases} \qquad (3)$$

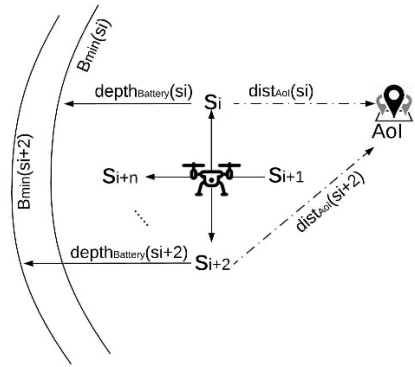

Figure 2: The structure of the Signed Distance in RDE domain

Then, we define a depth function to measure the distance between the position $s_i$ that is under robustness analysis and the UAV resource limit. We assume that the UAV starts its mission with full battery (B=100%) to explore the assigned environment. Given that the UAV moves with velocity $v$, we define a region centered at $s$ with radius $v \times B_{min}$ to find the farthest positions that the UAV could travel while still being able to go back home. With this region defined (Figure 2), we can define the function $depth_{battery}$.

**Definition 3** (Battery Life $depth$ function): Given that B is the current level of the battery life and $B_{min}$ is the battery minimum threshold, $depth_{battery}$ function for the UAV at position $s_i$ is defined as:

$$depth_{battery}(s_i) = \begin{cases} (B - B_{min}) - \frac{d(s_i, Home)}{v} & if\ s_i \notin \mathcal{O}(B_{min}) \\ 0 & otherwise \end{cases} \quad (4)$$

The function $depth_{battery}$ measures the distance to the closest edge of the region defined by a constraint centered on the position $s_i$. It should be noted that the defined regions include a third dimension for time. Therefore, the outer edges of the structure shown in Figure 2 would shrink over time.

Given a position $s_i$, we have defined a robustness metric $R_\varphi(s_i) = [\![(\varphi, \mathcal{O})]\!](s_i, t)$ that denotes how robustly $s_i$ satisfies (or falsifies) $\varphi$ at time t. The robustness metric $R_\varphi$ maps each position $s_i$ to a real number $\varepsilon$. The sign of $\varepsilon$ indicates whether $s_i$ satisfies $\varphi$ and its magnitude $|\varepsilon|$ measures its robustness value. More generally, given a robustness threshold $\rho > 0$ and a neighboring function $\zeta$ to return a set of positions which are in neighboring distance (i.e. within the range of the UAV) from the UAV's current location, we need to find:

$$s_i \in \zeta(s_{i-1})\ s.t.\ R_\varphi(s_i) \geq \rho \quad (5)$$

Using the $dist$ and $depth$ functions, the P-MTL robustness degree of $\varphi$ in equation 2 can be point-wise computed for each position $s_i$ under robustness analysis to solve the RDE problem in equation 5.

The robustness of the safety property in equation 2 measured at each neighbor position since it must hold during the whole trajectory. To measure the robustness of the safety constraint for position $s$, we use the MTL robustness semantic with duration of [1, 1] to guarantee the constraint satisfaction during all time steps. In order to apply the robustness semantic, the *always*, *eventually*, and *next* operators are converted into the *Until* operator using the conversion rules in [8]. Then, the robustness becomes a minimum function of the robustness of True value and the $depth_{battery}$ function as illustrated in equation 6. Since the robustness of True by semantic is positive infinity, the robustness function becomes about the value of $depth_{battery}$. Equation 6 measures how far away the UAV is from being out of battery if it chooses to explore position $s_i$.

$$\square(B(S_i) > B_{min})$$
$$= \neg(T\ \mathcal{U}\ B(S_i) > B_{min}) \overset{[1,1]}{\Longrightarrow} = \neg\left(\bigsqcup_{j=1}^{1} [\![T]\!](S_i, j) \sqcap \prod_{k=1}^{j} [\![B]\!](S_k, k)\right)$$
$$= \min(\infty, depth_{battery}(S_i))$$
$$= depth_{battery}(S_i) \quad (6)$$

The robustness of the liveness property evaluates the reachability of AoI from position $s_i$ in equation 7. The robustness becomes about the distance from $s_i$ to the closest AoI.

$$\lozenge\ p\left(inside(\bullet_{t|t}\ AoI)\right) > \beta$$
$$= (T\ \mathcal{U}\ AoI(S_i) > \beta)$$
$$\overset{[t,t]}{\Longrightarrow} = \left(\bigsqcup_{j=t}^{t} [\![T]\!](S_i, j) \sqcap \prod_{k=t}^{j} [\![AoI]\!](S_i, k)\right)$$
$$= \min(\infty, dist_{AoI}(S_i))$$
$$= dist_{AoI}(S_i) \quad (7)$$

On the other hand, the robustness of the conditional liveness property evaluates the ability of the UAV to avoid being stuck in non-AoI for longer than $v$ time steps in equation 8. This property forces the UAV to find another position closer to an AoI or to go to homebase and terminate the mission indicating that it has successfully explored the AoI of the given environment. The robustness of the P-MTL semantic for this property selects the closest position to an AoI. In order to be able to explore another AoI even when the neighbor positions are all classified as non-AoI, we develop a simple technique to allow the UAV to memorize the locations of previously seen but not-explored-yet areas that can be potentially classified as AoI inspired by the developed behavior of Frontier Exploration in [3]. We call those locations *cached points*. Hence, the UAV would keep a local list of cached points while exploring other areas with higher likelihood of being AoI in order to use them to satisfy its conditional liveness property.

$$\square p\left(inside(\bullet_{t-v|t}\ AoI)\right) < \lambda \Rightarrow \circ P\left(inside(\bullet_{t|t}\ AoI)\right)$$
$$> \lambda = \neg\left((T\ \mathcal{U}_{[t-v,t]}\ AoI(S_i) < \lambda) \wedge (T\ \mathcal{U}_{[t,t]}\ AoI(S_{i+1}) > \lambda)\right)$$
$$\overset{[t-v,t]}{\Longrightarrow} = \neg\begin{pmatrix}\left(\bigsqcup_{j=t-v}^{t} [\![T]\!](S_i, j) \sqcap \prod_{k=t-v}^{j} [\![AoI]\!](S_i, k) - \lambda\right) \\ \sqcap \left(\bigsqcup_{j=t}^{t} [\![T]\!](S_{i+1}, j) \sqcap \prod_{k=t}^{j} [\![AoI]\!](S_{i+1}, k) - \lambda\right)\end{pmatrix}$$
$$= \max(\min(\infty, dist_{AoI}(S_i)), \min(\infty, dist_{AoI}(S_{i+1})))$$
$$= \max(dist_{AoI}(S_i), dist_{AoI}(S_{i+1})) \quad (8)$$

The robustness function in equation 2 becomes about finding the minimum values of the results of equations 6-8.

### 4.1 MCMC Sampling

In this section, we explain a Markov Chain Monte Carlo sampling method that we use to solve equation 5 using the computed robustness in equations 6-8. The MCMC technique presented here is based on *acceptance-rejection* sampling [23]. Typically, Monte-Carlo based techniques are widely used for solving global optimization problems [24]. In this paper, we adopt a class of MCMC sampling techniques called the Metropolis-Hastings [23] to stochastically walk the UAV over a Markov chain that is defined by the P-MTL robustness.

**ALGORITHM 1: MCMC Sampling Algorithm**

**Input**: $s_i$: current position, $f(s_i) = R_\varphi(s_i)$ Robustness Function, $\rho$: Robustness threshold
**Output**: $s_{i+1} \in \zeta(s_i) \cup s_i$
**begin**
 Uniformly choose one random neighbor $s' \in \zeta(s_i)$
**if** $f(s') > \rho$ && $f(s') > f(s_i)$
    return $s'$
**else**
    $\sigma = e^{-\left(\tau(f(s') - f(s_i))\right)}$
    r ← UniformRandomReal(0, 1) ;
    **if** (r ≤ σ) **then**
        return $s'$
    **else**
        return $s_i$
**end**

Our sample space consists of the neighbors of the UAV's current position such that the next generated position for the UAV to explore is randomly selected satisfying the problem specification in equation 5. Algorithm 1 maximizes the robustness of equation 5 to find a position that has higher estimation of AoI. First, the function $\zeta$ is used to find the neighbors of the input position $s_i$. Then, the algorithm uniformly chooses one random neighbor $s'$ and sample the robustness function at the neighbor $f(s')$. If $f(s') > f(s_i)$, then the neighbor position is returned as the next target. Otherwise, the ratio $\sigma = \exp\left(-\left(\tau(f(s') - f(s_i))\right)\right)$ is computed as the acceptance probability for the new proposal. Note that if $\sigma \geq 1$ (i.e, $f(s') \geq f(s_i)$), then the proposed neighbor is accepted with certainty. Even if $f(s') < f(s_i)$ the proposal may still be accepted with some non-zero probability. If the proposal is accepted, then $s'$ is returned as the next target position. Failing this, $s_i$ remains as the next target. In general, MCMC techniques require the design of a *proposal scheme* for choosing a proposal $s'$ given the current position $s_i$. The convergence of the sampling to the underlying distribution defined by $f$, depends critically on the choice of this proposal distribution. In this paper, we choose the Gibbs-Boltzmann function following the Metropolis-Hastings algorithm [23] because of its relatively fast convergence. In Gibbs-Boltzmann distribution, $\tau$ is a constant $1/kT$, which is the inverse of the product of Boltzmann's constant k and thermodynamic temperature $T$.

### 4.2 RDE Algorithm

Algorithm 2 implements a local-search technique in an unknown environment to compute a trajectory that would lead the UAV to navigate more AoI while maintaining its battery constraint. The algorithm starts by picking a random position to begin the flight. The algorithm would move the UAV at each time step to a position with a robustness larger than $\rho$ generated via the MCMC algorithm (Algorithm1). However, MCMC is a stochastic algorithm by nature and it could take many iterations to converge from the current position to the target position with an acceptable robustness. Moreover, MCMC runs the risk of getting stuck in local maxima;

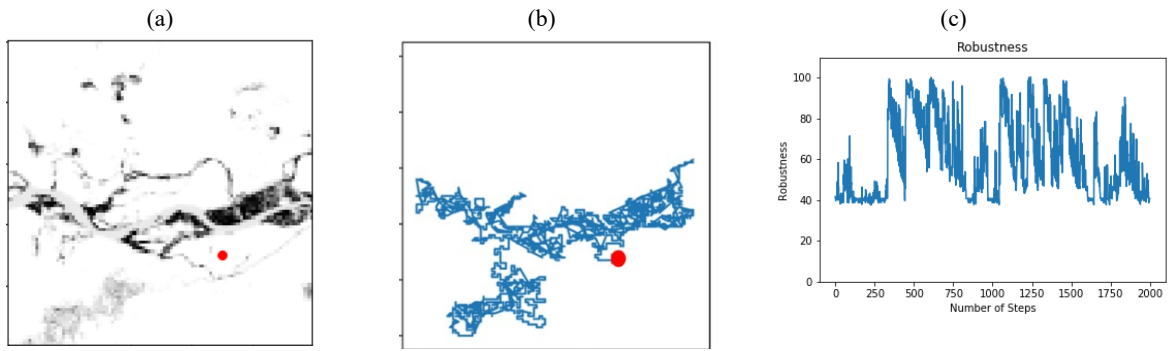

Figure 3: (a) Satellite image from Amazon Forest in Peru, (b) Flight trajectory generated by RDE, and (c) robustness value of the exploration decision at each time step.

areas where the robustness is higher for the current position than for its close neighbors, but lower than for locations that are further away. This could potentially happen when the UAV explores a large area with little to zero significant interest. This is remedied by setting a threshold α to stop the MCMC from generating the same results and enforce the algorithm to use one of the cached points, which in this case, represent further away locations with more robustness values. After making a decision about the next target, we use RA*, a path planner algorithm that has been developed using MTL robustness and A* [10] to find the path from the current to the next positions that would give the UAV exposure to more AoI if there is any around the path.

Back to ASGM example, Figure 3(a) shows a simulated map of the likelihood of finding ASGM for an area in Amazon forest in Peru. The darker the color the higher the likelihood is for the area to have ASGM. Such likelihood values would be provided by the object detection system onboard the UAV for small areas within its range of vision. The red circle represents the starting point of the flight. Figure 3(b) shows the flight trajectory that satisfies our RDE specification in equation 2 and generated by Algorithm 2 such that AoI is defined as areas of ASGM. Figure 3(c) plots the robustness of the exploration decision at each time step. Clearly, the selected positions for the UAV's trajectory in the given map are concentrated in the more promising regions with higher robustness values above $\rho = 38$. However, the resulting trajectory directly depends on the starting point and the number of steps which simulates the battery life of the UAV. More details about this experiment are shown in next section.

---

**ALGORITHM 2: RDEAlgorithm**

**Input**: $\varphi$ (2): Mission specification, $f(.) = R_\varphi(.)$: Robustness Function, $\rho$: Robustness threshold, $\zeta(.)$: neighboring function.
**begin**
 Randomly pick a starting point $s_0$
 While ($B > B_{min}$)
     count=0
     While $s_i == s_{i-1}$ && count< $\alpha$
         $s_i$ =MCMC($s_{i-1}$, f($s_{i-1}$), $\rho$)
         count++
     end
     if count>$\alpha$ && cachedPoints!=$\phi$
         $s_i$ = getCachedPoint();
     else
         $s_i$ = home;
     RA*($s_{i-1}, s_i$)
**End**

---

## 5 EXPERIMENTS

The main motivation for this paper is to increase the UAV's exploration percentage of AoI in unknown environments when the UAV has limited resources (i.e. battery and onboard storage) to completely explore the whole region. We implemented our approach to navigate a region within the Amazon forest in Peru to look for illegal ASGM as described in Example (ASGM).

As part of this research, we have developed an object recognition module using YOLO [25]. Our object recognition has been trained to detect different components that are usually found around ASGM areas such as dredges, floats, sluices, shacks/rooftops, sand, water, and plantations. The results from the object recognition have been simulated in this paper to test the proposed RDE approach. An example of the detection results of mining sites is shown in Figure 4.

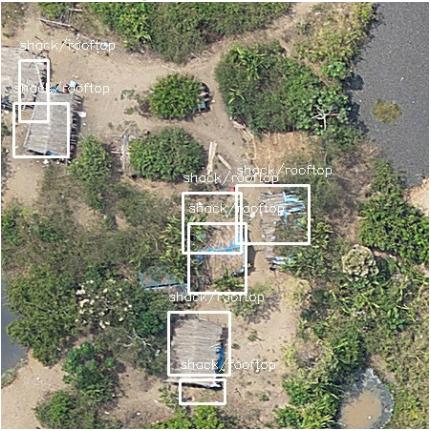

Figure 4: An example of the detection results generated by our object recognition module

To help guide the UAV flights to areas with real information, we implemented our RDE approach over actual 8x8 km$^2$ region containing ASGM (Figure 5 and Figure 6). We simulate motion of the UAV (as well as the onboard object detection system) and keep its altitude fixed by setting the field of view to 200x200 m$^2$. We test our RDE approach against the AEP approach developed in [3]. However, due to the space limitation, we only included the results from two maps that are shown in Figure 5. The focus of AEP is on combining RRT algorithm and Frontier Exploration [2] to explore the whole region without focusing on specific areas as AoI. We modified AEP to consider exploring the ASGM areas by changing their cost function to prioritize the areas with higher likelihood of having mining sites.

Figure 6 (a) shows the likelihood of ASGM areas in the region of shown in (Figure 5), the color scale is between yellow and green such that dark yellow areas have higher probability of having ASGM. Figure 6 (b) shows the most frequent explored positions using the proposed RDE approach. We collected those points by running RDE on 100 trials with 2000 time steps per each trail starting from random positions in each run. The green and yellow colors represent the most visited areas such that areas in yellow are visited more than areas with green color. We then explored the same map using the modified AEP (Figure 6 (c)). The testing results for the map shown in Figure 5 (b) are illustrated in Figure 7. For both maps, our approach was clearly able to navigate the majority of ASGM areas in comparison to AEP while spending less time inside vegetation areas. However, AEP was faster in making decisions than RDE by 34% and 24% when exploring the maps in Figure 5 a and b respectively. AEP uses a greedy algorithm which guaranteed faster execution but not necessarily good coverage for AoI while RDE needs to compute the robustness of P-MTL constraints before each exploration decision and use the MCMC sampler to select the next target with higher robustness.

Figure 8 illustrates the exploration of ASGM in the map shown in Figure 5.a using our RDE and AEP with different numbers of time steps respectively. The time steps here represent the battery life for the UAV. The percentage of coverage grows linearly with the allotted time for both approaches, but the RDE covers more ASGM areas by approximately 38% over AEP.

## 6 CONCLUSION

In this paper, we presented a new exploration approach RDE that incorporates the online discovered knowledge into the exploration decisions for UAVs. RDE uses the robustness of P-MTL specifications to guide the stochastic process of MCMC to make the exploration decisions in completely unknown environment. We have tested our approach on four simulated maps in Amazon forest in Peru to look for mining areas (e.g. ASGM). In comparison to a greedy approach called AEP [], our approach leads the UAV into more areas classified as ASGM than AEP without getting stuck or spending long time in vegetation areas. In future work, we intent to test our approach on real UAVs in Amazon forest. In order to do that, we have to incorporate the dynamics of the UAV and the control information (i.e. speed, altitude) into the P-MTL specifications of the problem. Moreover, we plan to improve the accuracy of our object recognition module by developing a data fusion technique to integrate low resolution data from satellite images with the UAV imagery for ASGM sites.

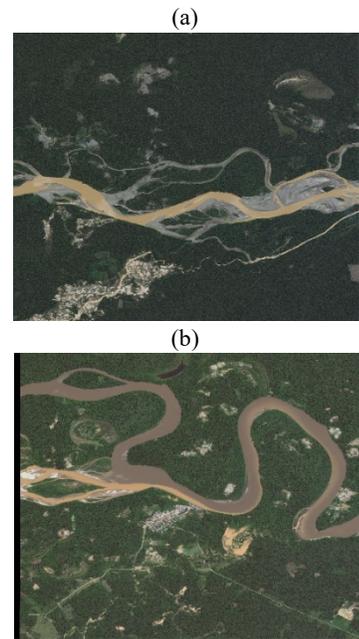

Figure 5: Satellite images for (a) area of NW/SE: -69.933, -12.724, -69,860, -12.79 (b) area of NW/SE: -70.425, -12.577, -70.351, -12.650

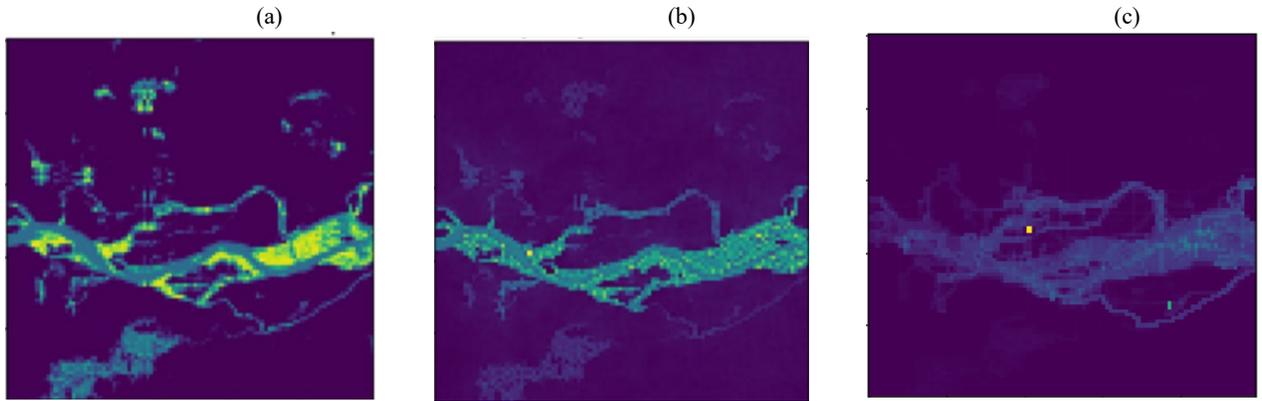

Figure 6: (a) Distributed ASGM in the map (Figure 5.a), (b) Distribution of most frequent explored locations using RDE, and (c) Distribution of most frequent explored locations using AEP

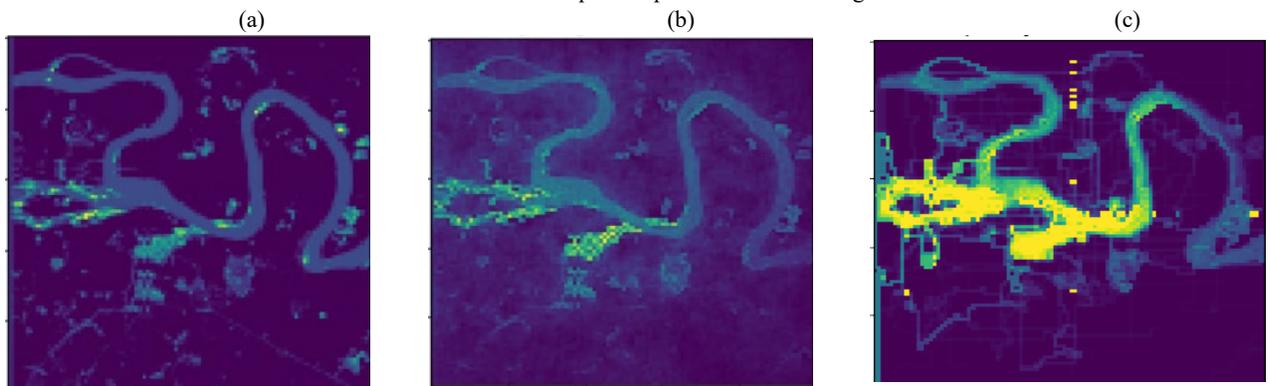

Figure 7: (a) Distributed ASGM in the map (Figure 5.b), (b) Distribution of most frequent explored locations using RDE, and (c) Distribution of most frequent explored locations using AEP

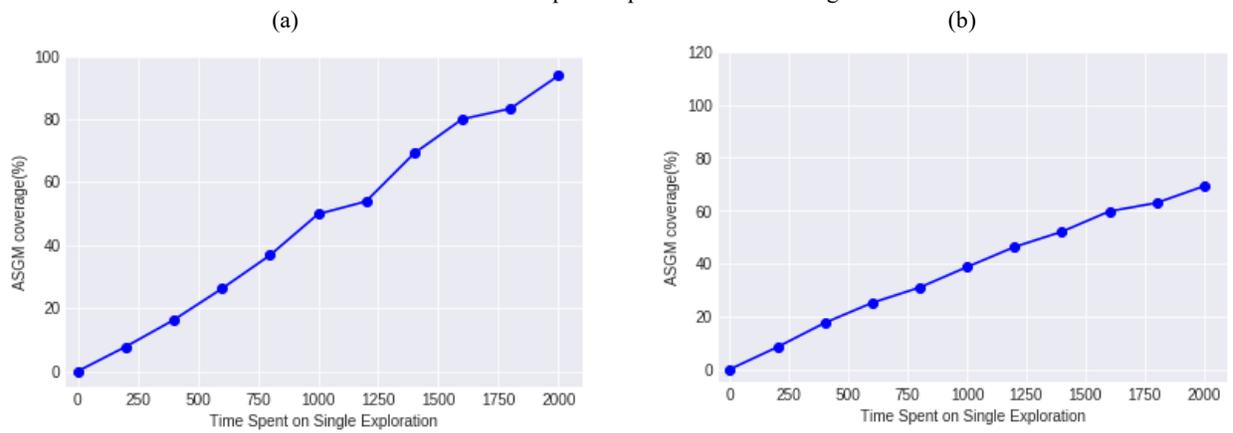

Figure 8: (a) ASGM coverage with RDE (b) ASGM coverage with AEP